\documentclass[conference]{IEEEtran}
\IEEEoverridecommandlockouts
\usepackage{comment}
\usepackage{cite}
\usepackage{amsmath,amssymb,amsfonts}
\usepackage{algorithmic}
\usepackage{graphicx}
\usepackage{textcomp}
\usepackage[caption=false,font=footnotesize]{subfig}
\usepackage{booktabs}
\usepackage[table]{xcolor}

\definecolor{faintgreen}{RGB}{220,245,220}

\newcommand{\hi}[1]{%
  \begingroup
  \setlength{\fboxsep}{1pt}%
  \colorbox{faintgreen}{\strut #1}%
  \endgroup
}

\usepackage{tabularx}
\usepackage{array}
\usepackage{multirow}
\usepackage{url}

\begin{document}

\title{Models in the Same Family are NOT Trust-Equivalent}

\author{
\begin{minipage}{\textwidth}
\centering

\begin{minipage}{0.30\textwidth}
\centering
Rohit Raj Rai\\
\textit{Department of Computer Science}\\
\textit{and Engineering}\\
\textit{Indian Institute of Technology}\\
\textit{Guwahati}\\
Guwahati, Assam, India\\
rohitraj@iitg.ac.in
\end{minipage}
\hfill
\begin{minipage}{0.30\textwidth}
\centering
Chirag Kothari\\
\textit{Department of Electrical}\\
\textit{Engineering}\\
\textit{Indian Institute of Technology}\\
\textit{Indore}\\
Indore, Madhya Pradesh, India\\
chiragkothari2503@gmail.com
\end{minipage}
\hfill
\begin{minipage}{0.30\textwidth}
\centering
Siddhesh Shelke\\
\textit{Department of Metallurgical}\\
\textit{Engineering and Materials}\\
\textit{Science}\\
\textit{Indian Institute of Technology}\\
\textit{Indore}\\
Indore, Madhya Pradesh, India\\
siddheshshelke015@gmail.com
\end{minipage}

\vspace{0.5cm}

\begin{minipage}{0.30\textwidth}
\centering
Yatika Jena\\
\textit{Department of Electronics and}\\
\textit{Communication Engineering}\\
\textit{Indian Institute of Technology}\\
\textit{Guwahati}\\
Guwahati, Assam, India\\
j.yatika@iitg.ac.in
\end{minipage}
\hspace{1.5cm}
\begin{minipage}{0.30\textwidth}
\centering
Amit Awekar\\
\textit{Department of Computer Science}\\
\textit{and Engineering}\\
\textit{Indian Institute of Technology}\\
\textit{Guwahati}\\
Guwahati, Assam, India\\
awekar@iitg.ac.in
\end{minipage}

\end{minipage}
}

\maketitle

\begin{abstract}
Within a model family, a smaller variant is often deployed as a drop-in replacement for a larger one when their performance is similar. However, performance alone does not tell the full story. We propose a framework to evaluate trust-equivalence between a larger model and a smaller one in the same family along two dimensions. The first is attribution alignment: do both models base their predictions on the same input features? The second is calibration similarity: do both models share the same relationship between confidence and accuracy?
We evaluate the Llama-2 family on two text classification tasks: Natural Language Inference and Paraphrase Identification. Attribution alignment is measured using two well-known methods: LIME and SHAP. Agreement between model pairs is quantified via the Jaccard coefficient over top-K attributed features. We observe that attribution alignment between models is generally low, indicating that smaller and larger models base their predictions on different input features. Calibration similarity is assessed using ECE, MCE, Brier Score, and Reliability Diagrams. Calibration profiles differ substantially across model sizes. There is no consistent relationship between model size and calibration quality. We have additionally verified these trends on two encoder-only families: BERT and Vision Transformer. The results are consistent with those reported here.
Our experimental results show that replacing a larger model with a smaller one from the same family is a multidimensional decision that requires consideration beyond performance measures alone. Trust-equivalence must be assessed explicitly. It cannot be assumed from performance alone.
\end{abstract}

\begin{IEEEkeywords}
Large language models, Interpretability, Calibration, Trust-equivalence, Transformer models
\end{IEEEkeywords}

\section{Introduction}
Smaller variants of Large Language Models (LLMs) are increasingly used as drop-in replacements for their larger counterparts. They offer lower
latency, smaller memory footprint, and reduced compute cost. Moreover, when a smaller model achieves comparable accuracy, the replacement case seems straightforward. Please refer to Table~\ref{tab:llama_model_family} for an illustration. The decision appears to be a simple efficiency trade-off with no downside.

However, accuracy is only one dimension of model behavior. Existing work on model selection within a family focuses almost entirely on performance metrics. This narrow focus leaves a silent assumption unchallenged: a model named ``Llama-2-7B'' behaves like ``Llama-2-70B'', just smaller and faster. Other aspects of model behavior, such as how it arrives at its predictions and how reliable its confidence is, receive little attention during evaluation.

Two examples illustrate what is at stake. Consider a doctor who relies on high-confidence predictions from a large diagnostic model. A smaller replacement matches its overall accuracy. However, its accuracy is poor, specifically for high-confidence predictions. Deploying such a model in a clinical setting could be fatal. Now consider a loan approval system. A smaller model replaces a larger one with matching accuracy. But the smaller model bases its decisions on sensitive attributes such as gender and religion. This feature attribution can lead to unfair outcomes and legal repercussions. Both examples make the same point. Performance parity is necessary but not sufficient for safe deployment.

We propose a framework for evaluating trust equivalence between a larger LLM and a smaller one within the same family. Trust-equivalence has two dimensions. The first is attribution alignment: do both models base their predictions on the same input features? In a Paraphrase Identification (PI) task, for instance, both models should assign high importance to the same words. The second is calibration similarity: for predictions made at the same confidence level, do both models achieve the same accuracy? For a target task, if we are satisfied with a large model's attribution and calibration profile, the smaller replacement should demonstrate the same. Importantly, our framework does not judge which model is better. It only checks whether the two models can be trusted in the same way.

We evaluate our framework empirically on the Llama-2 family, a decoder-only LLM with 7B, 13B, and 70B variants. They were each pretrained independently from scratch on the same corpus. We use two text
classification tasks: Natural Language Inference using the SNLI dataset and Paraphrase Identification using the QQP dataset. Attribution alignment is measured using LIME and SHAP. Calibration similarity is assessed using ECE, MCE, Brier Score, and Reliability Diagrams. We additionally validate our findings on the encoder-only BERT family and the Vision Transformer (ViT) family. The trends are consistent across all three families. In this paper, we discuss Llama-2 results in detail. BERT and ViT results are discussed
briefly due to space constraints. Detailed results for BERT and ViT families are available along with our code repository.

Our results show that trust-equivalence cannot be assumed from performance parity alone. Attribution alignment between a larger and a smaller Llama-2 model is generally low. It indicates that the two models base their predictions on different input features. Calibration profiles differ substantially across model sizes, with no consistent
relationship between size and calibration quality. These findings hold across encoder-only, decoder-only, and vision transformer families. Replacing a larger model with a smaller one is not just
an efficiency decision. It is also a trust decision, and it requires explicit measurement.

This paper makes the following contributions. First, we propose a two-dimensional trust-equivalence framework applicable to any pair of models in the same family. Second, we provide a comprehensive empirical evaluation across two tasks and multiple evaluation tools. Third, we show that performance parity does not imply trust-equivalence. In summary, our work makes a case for moving beyond performance-centric evaluation when a larger LLM is replaced by a smaller one in the same family. For complete reproducibility, all our code, datasets, trained models, and additional results on BERT and ViT families are available publicly on the Web\footnote{https://github.com/anonymouscode816/small-models-trust-equivalence}.

\begin{table}[hbt]
\centering
\caption{Smaller versions of Llama-2 offer slight reduction in accuracy with significant efficiency gain.}
\label{tab:llama_model_family}
\scriptsize
\setlength{\tabcolsep}{6pt}
\renewcommand{\arraystretch}{1.15}
\begin{tabular}{lcccc}
\toprule
\textbf{Model} &
\textbf{\shortstack{Parameters\\(Billions)}} &
\textbf{\shortstack{Size\\(GB)}} &
\textbf{\shortstack{NLI\\Accuracy}} &
\textbf{\shortstack{PI\\Accuracy}} \\
\midrule
Llama-2-70B & 70 & 137.98 & 91.10 & 92.30 \\
Llama-2-13B & 13 & 26.03  & 88.34 & 89.10 \\
Llama-2-7B  & 7  & 13.48  & 85.49 & 89.00 \\
\bottomrule
\end{tabular}
\end{table}

\section{Related Work}\label{sec2}
Model families that span a range of sizes are now standard practice
in deep learning. The goal is to give practitioners a size variant
that fits their resource budget~\cite{gupta2022compression,
li2023model}. With the dominance of the Transformer
architecture~\cite{vaswani2017attention}, families like
Llama-2~\cite{touvron2023llama} are released with variants trained
independently from scratch on the same corpus. Across all such
families, size variants are evaluated almost exclusively on accuracy.
Efficiency is the objective. Trust is never part of the evaluation.

Researchers have studied trustworthiness of individual models across
several dimensions, but these efforts developed in isolation from
each other. As deep learning moved into critical domains like
healthcare and finance~\cite{esteva2017dermatologist,
rajpurkar2017chexnet}, AI safety researchers showed that harmful
behavior can emerge from poor objective design and distributional
shift~\cite{amodei2016concrete}. Explainability researchers developed
methods like LIME~\cite{ribeiro-etal-2016-trust} and
SHAP~\cite{NIPS2017_8a20a862} to open the black
box~\cite{guidotti2018survey}. Bias researchers showed that models
discriminate based on sensitive attributes in criminal
sentencing~\cite{angwin2016machine} and word
embeddings~\cite{bolukbasi2016man}. Calibration researchers showed
that model confidence is poorly aligned with actual
accuracy~\cite{guo2017calibration}. Each line of work is important.
But none of them asked what happens to these properties when a
larger model is replaced by a smaller one.

Prior work on the behavioral effects of size reduction focused on
robustness and fairness, not on trust-equivalence as a pairwise
property. Hooker et al. showed that smaller models forget certain
subgroups disproportionately~\cite{hooker2019compressed}. Ye et al.
showed that size reduction degrades robustness to adversarial
attacks~\cite{ye2019adversarial}. Yuan and Zhang showed that smaller
models are more vulnerable to membership inference
attacks~\cite{yuan2022membership}. Rai et al. showed that smaller
models are not miniature versions of their large counterparts across
data representation, prediction errors, and adversarial
vulnerability~\cite{rai2024compressed}. These findings establish
that size matters beyond accuracy. But they do not ask whether a
smaller model can be trusted in the same way as its larger
counterpart.

The most relevant recent work evaluates trustworthiness of LLMs,
but treats each model in isolation. Hong et al. studied
trustworthiness of smaller LLMs across privacy, fairness, and
robustness~\cite{pmlr-v235-hong24a}. Wang et al. evaluated
trustworthiness of GPT models
comprehensively~\cite{wang2023decodingtrust}. These are valuable
contributions. But they ask whether a model is trustworthy in an
absolute sense. They do not ask whether a smaller model and a
larger model in the same family can be trusted equivalently. A
deployment engineer does not ask the absolute question. They ask:
is this smaller model a trust-equivalent replacement for the larger
one I have already validated? This is the question our work
addresses.

\subsection{Evaluation Metrics}

We use two categories of metrics in our framework.
The first category measures attribution alignment.
The second measures calibration similarity.

To measure attribution alignment, we use the Jaccard Coefficient~\cite{jaccard1912distribution}. It measures the overlap between two sets. Given two sets $A$ and $B$, it is defined as:

\begin{equation}
J(A, B) = \frac{|A \cap B|}{|A \cup B|} =
\frac{|A \cap B|}{|A| + |B| - |A \cap B|}
\end{equation}

A score of $1$ means perfect agreement. A score of $0$ means no overlap.
We apply it to the sets of top-$K$ features selected by each model.
It is widely used in information retrieval to compare binary attributes and token sets.

To measure calibration similarity, we use three complementary metrics and reliability diagrams. Each captures a different aspect of the alignment between model confidence and actual accuracy.

Expected Calibration Error (ECE)~\cite{guo2017calibration} partitions predictions into $M$ equally spaced bins ($B_1$ to $B_M$) in terms of confidence. Out of total $N$ predictions, number of predictions in each partition $B_m$ is $|B_m|$. For each bin, the mean confidence ($conf(B_m)$) and the observed accuracy ($acc(B_m)$) are computed. ECE computes the weighted average of the absolute difference between observed accuracy and mean confidence across bins:

\begin{equation}
ECE = \sum_{m=1}^{M} \frac{|B_m|}{N}
\left| acc(B_m) - conf(B_m) \right|
\end{equation}

Maximum Calibration Error (MCE) captures the worst-case deviation between accuracy and confidence across all bins. It is especially relevant for high-stakes applications:

\begin{equation}
MCE = \max_{m \in \{1, \dots, M\}}
\left| acc(B_m) - conf(B_m) \right|
\end{equation}

The Brier Score~\cite{glenn1950verification} measures the mean squared difference between predicted probabilities and actual outcomes. It penalizes predictions where the correct label's probability deviates from $1$:

\begin{equation}
BS = \frac{1}{N} \sum_{i=1}^{N} \sum_{j=1}^{C} (f_{ij} - o_{ij})^2
\end{equation}

where $f_{ij}$ is the predicted probability for class $j$ on instance $i$, and $o_{ij} \in \{0, 1\}$ is the indicator of whether class $j$ is the true label for instance $i$.

For all three metrics, a lower value indicates better calibration. They are complementary rather than interchangeable. ECE measures average miscalibration. MCE measures worst-case miscalibration. The Brier Score additionally penalizes lack of sharpness, that is the failure to assign high probability to correct predictions. No single metric captures the full picture.

Reliability Diagrams provide a visual complement to the numerical metrics above. Originally proposed by DeGroot and Fienberg~\cite{degroot1983comparison} and later popularized by Niculescu-Mizil and Caruana~\cite{niculescu2005predicting}, they have become a standard diagnostic tool for model calibration. Predictions are grouped into $M$ bins based on their confidence scores. For each bin, the mean confidence and the observed accuracy are computed. The diagram plots observed accuracy on the Y-axis against mean confidence on the X-axis. A perfectly calibrated model follows the identity line $y = x$. A curve below the diagonal indicates overconfidence; a curve above indicates underconfidence. The density of predictions in each bin also matters. A bin containing very few predictions produces an unreliable estimate of accuracy~\cite{vaicenavicius2019evaluating}. This is why we report confidence bucket distributions alongside our reliability diagrams.

\section{Trust-Equivalence Framework}

We define a model $M_1$ to be \textit{trust-equivalent} to a model $M_2$ if two conditions hold. First, $M_1$ generates predictions through the same decision process as $M_2$. Second, $M_1$ demonstrates the same probability reliability as $M_2$. We refer to the first condition as \textit{attribution alignment} and the second as \textit{calibration similarity}.

attribution alignment asks whether both models base their predictions on the same input features. Consider the PI task. A larger model and a smaller one may both predict correctly that a given pair of sentences are near duplicates. But they may focus on entirely different words to reach that prediction. Their predictions agree. Their decision processes do not. The same issue arises in the NLI task. Two models may assign the same label to a premise-hypothesis pair. But one may focus on negation cues while the other focuses on unrelated words. Accuracy-based evaluation misses this difference entirely.

Calibration similarity asks whether, for predictions made at the same confidence level, both models achieve the same accuracy. Consider a larger model that is correct 90\% of the time when it predicts with 90\% confidence. If its smaller replacement is correct only 70\% of the time at the same confidence level, a user relying on that confidence score will be misled. The two models are not calibration-similar. Their overall accuracy may still be identical.

Our framework does not judge which model is better. It does not ask whether $M_1$ is more accurate or better calibrated than $M_2$ in an absolute sense. It asks only whether they can be trusted in the same way. A deployment engineer who has validated a larger model for a critical application needs to know whether a smaller replacement will behave equivalently. Overall performance metrics do not answer this question. Trust-equivalence does.

Please refer to Table~\ref{tab:llama_model_family}. It reports accuracy and inference latency across our evaluation tasks. A consistent pattern emerges. Llama-2-7B is ten times smaller than Llama-2-70B yet remains competitive on both NLI and PI. This is precisely the scenario where the temptation to deploy a smaller model as a drop-in replacement is strongest. It is also precisely where trust-equivalence must be assessed before making that choice.

\subsection{Model Family}
\label{sec:model_families}

Llama-2~\cite{touvron2023llama} is a decoder-only transformer family released by Meta. We evaluate three variants: Llama-2-7B, Llama-2-13B, and Llama-2-70B. Each variant was trained independently from scratch on two trillion tokens. The smaller variants are not derived from the larger one. They are distinct models with their own weight initializations. 


We chose Llama-2 over more recent Llama versions for a principled reason. Our study requires a family with at least three independently trained size variants. Llama-3 and Llama-3.1 offer only two variants runnable on a single GPU, as their 405B model requires multi-node infrastructure. Llama-3.2 has only two text-only variants (1B, 3B). Mixing variants across sub-versions would introduce training data and protocol inconsistencies that would confound our trust-equivalence measurements. Llama-4 Scout and Maverick both have 17B active parameters, offering no meaningful variation in active model size. Their
total parameter counts differ due to different numbers of experts, but this reflects architectural variation rather
than scale. Llama-2's 7B, 13B, and 70B variants are trained under an identical protocol on the same corpus, all runnable on our single NVIDIA A100 GPU, and span a 10x parameter range. These characteristics make it the only decoder-only Llama family satisfying our requirements cleanly. It is also well documented, widely used, and fully reproducible via the Hugging Face library.\footnote{\url{https://huggingface.co/meta-llama/Llama-2-7b-hf}}
\footnote{\url{https://huggingface.co/meta-llama/Llama-2-13b-hf}}
\footnote{\url{https://huggingface.co/meta-llama/Llama-2-70b-hf}}

Several other open-weight LLM families exist apart from Llama. OPT~\cite{zhang2022opt} and Pythia~\cite{biderman2023pythia} offer
controlled multi-size variants but predate Llama-2 and are less representative of current deployment practice. Mistral lacks a clean family of dense variants at multiple scales. Phi-3 offers multiple size variants within a single generation, but its largest variant is only 14B, limiting the parameter range to approximately 4x. Llama-2 spans a 10x range with a 70B anchor that is more representative of large model deployment scenarios. Qwen offers comparable multi-size variants on a shared corpus, but its pretraining data composition is less transparently documented than Llama-2's. that makes it harder to isolate size as the sole variable. We choose Llama-2 because
it combines transparent and documented pretraining, a meaningful 10x parameter range across three variants, and wide adoption in
production deployment settings.
\begin{table*}[hbt]
\caption{attribution alignment Example using the LIME test (Task: NLI, Model family: Llama-2).
For this datapoint, the ground truth label is ``Neutral''.
All models correctly predict the ground truth label.
For each model, we highlight the top 3 important words.}
\label{tab_example_alignment_llama_NLI}
\centering
\scriptsize
\renewcommand{\arraystretch}{1.15}
\setlength{\tabcolsep}{4pt}
\begin{tabularx}{\textwidth}{@{}>{\raggedright\arraybackslash}X
                                >{\raggedright\arraybackslash}X
                                >{\raggedright\arraybackslash}p{0.20\textwidth}@{}}
\toprule
\textbf{Premise} & \textbf{Hypothesis} & \textbf{Model} \\
\midrule

\hi{A} long jumper takes flight in an attempt to jump the longest distance possible in a track competition.
&
A long jumper tries to beat the person \hi{before} him, but he hasn't performed well \hi{recently}.
&
70B \\

\addlinespace

A long jumper takes flight in an attempt to jump the longest distance possible in a track competition.
&
A long jumper \hi{tries} to beat the \hi{person} \hi{before} him, but he hasn't performed well recently.
&
13B \\

\addlinespace

A long jumper \hi{takes} flight in an \hi{attempt} to jump the longest distance possible in a track competition.
&
A long jumper tries to beat the person before him, but he hasn't \hi{performed} well recently.
&
7B \\

\hline
\end{tabularx}
\end{table*}

\begin{table*}[hbt]
\caption{attribution alignment Example using the LIME test (Task: PI, Model family: Llama-2).
For this datapoint, the ground truth label is ``Not Duplicate''.
All models correctly predict the ground truth label.
For each model, we highlight the top 3 important words.}
\label{tab_example_alignment_llama_PI}
\centering
\scriptsize
\renewcommand{\arraystretch}{1.15}
\setlength{\tabcolsep}{4pt}
\begin{tabularx}{\textwidth}{@{}>{\raggedright\arraybackslash}X
                                >{\raggedright\arraybackslash}X
                                >{\raggedright\arraybackslash}p{0.20\textwidth}@{}}
\toprule
\textbf{Question-1} & \textbf{Question-2} & \textbf{Model} \\
\midrule

I often \hi{cry} for small little things, though I really wanted to be strong in front of others I couldn't change \hi{this} behaviour. how to come out of this?
&
What \hi{are} excavation hazards, and what are some examples?
&
70B \\

\addlinespace

I often cry for \hi{small} little things, though I really wanted to be strong in front of others I couldn't change this behaviour. \hi{how} to come out of this?
&
What are \hi{excavation} hazards, and what are some examples?
&
13B \\

\addlinespace

I often cry for small little things, \hi{though} I really wanted to be \hi{strong} in front \hi{of} others I couldn't change this behaviour. how to come out of this?
&
What are excavation hazards, and what are some examples?
&
7B \\

\hline
\end{tabularx}
\end{table*}






\begin{table}[hbt]
\caption{attribution alignment between Llama-2 models using Jaccard coefficient ($K=3$).}
\label{tab_alignment_results_llama}
\centering
\scriptsize
\setlength{\tabcolsep}{3pt}
\renewcommand{\arraystretch}{1.1}

\begin{tabular}{llcccc}
\toprule
\multirow{2}{*}{$M_1$} &
\multirow{2}{*}{$M_2$} &
\multicolumn{2}{c}{\textbf{NLI}} &
\multicolumn{2}{c}{\textbf{PI}} \\
\cmidrule(lr){3-4}
\cmidrule(lr){5-6}
& & \textbf{LIME} & \textbf{SHAP}
& \textbf{LIME} & \textbf{SHAP} \\
\midrule
70B & 13B & 0.205 & 0.574 & 0.185 & 0.388 \\
70B & 7B  & 0.193 & 0.344 & 0.177 & 0.261 \\
\midrule

13B & 7B & 0.180 & 0.358 & 0.172 & 0.317 \\
\bottomrule

\end{tabular}
\end{table}

\begin{table}[hbt]
\renewcommand{\arraystretch}{1.1}
\caption{Softmax (S) and verbalized (V) confidence distributions for NLI.}
\label{tab_llama_nli_calibration_bins}
\centering
\scriptsize
\setlength{\tabcolsep}{2pt}

\begin{tabular}{lcccccc}
\toprule
\textbf{Bin}
& \multicolumn{2}{c}{\textbf{70B}}
& \multicolumn{2}{c}{\textbf{13B}}
& \multicolumn{2}{c}{\textbf{7B}} \\
\cmidrule(lr){2-3}
\cmidrule(lr){4-5}
\cmidrule(lr){6-7}
& \textbf{S} & \textbf{V}
& \textbf{S} & \textbf{V}
& \textbf{S} & \textbf{V} \\
\midrule
0.0--0.1 & 0.00 & 0.00 & 0.00 & 0.06 & 0.00 & 0.00 \\
0.1--0.2 & 0.00 & 0.00 & 0.00 & 0.00 & 0.00 & 0.00 \\
0.2--0.3 & 0.00 & 0.00 & 0.00 & 0.00 & 0.00 & 0.00 \\
0.3--0.4 & 0.57 & 0.00 & 0.35 & 0.00 & 0.43 & 0.00 \\
0.4--0.5 & 4.41 & 0.13 & 3.37 & 0.00 & 2.98 & 0.00 \\
0.5--0.6 & 8.03 & 0.00 & 7.96 & 0.00 & 7.94 & 0.00 \\
0.6--0.7 & 11.91 & 4.95 & 9.64 & 0.00 & 8.71 & 0.00 \\
0.7--0.8 & 19.82 & 13.67 & 14.60 & 0.30 & 10.37 & 0.00 \\
0.8--0.9 & 22.83 & 28.17 & 21.19 & 0.00 & 16.17 & 0.00 \\
0.9--1.0 & 32.43 & 53.08 & 42.89 & 99.64 & 53.39 & 100.00 \\
\midrule
\textbf{Avg.}
& \textbf{80.18} & \textbf{89.82}
& \textbf{83.23} & \textbf{98.73}
& \textbf{84.86} & \textbf{99.00} \\
\bottomrule
\end{tabular}
\end{table}


\begin{figure}[hbt]
\centering

\begin{minipage}[t]{0.48\columnwidth}
\centering
\includegraphics[width=\linewidth]{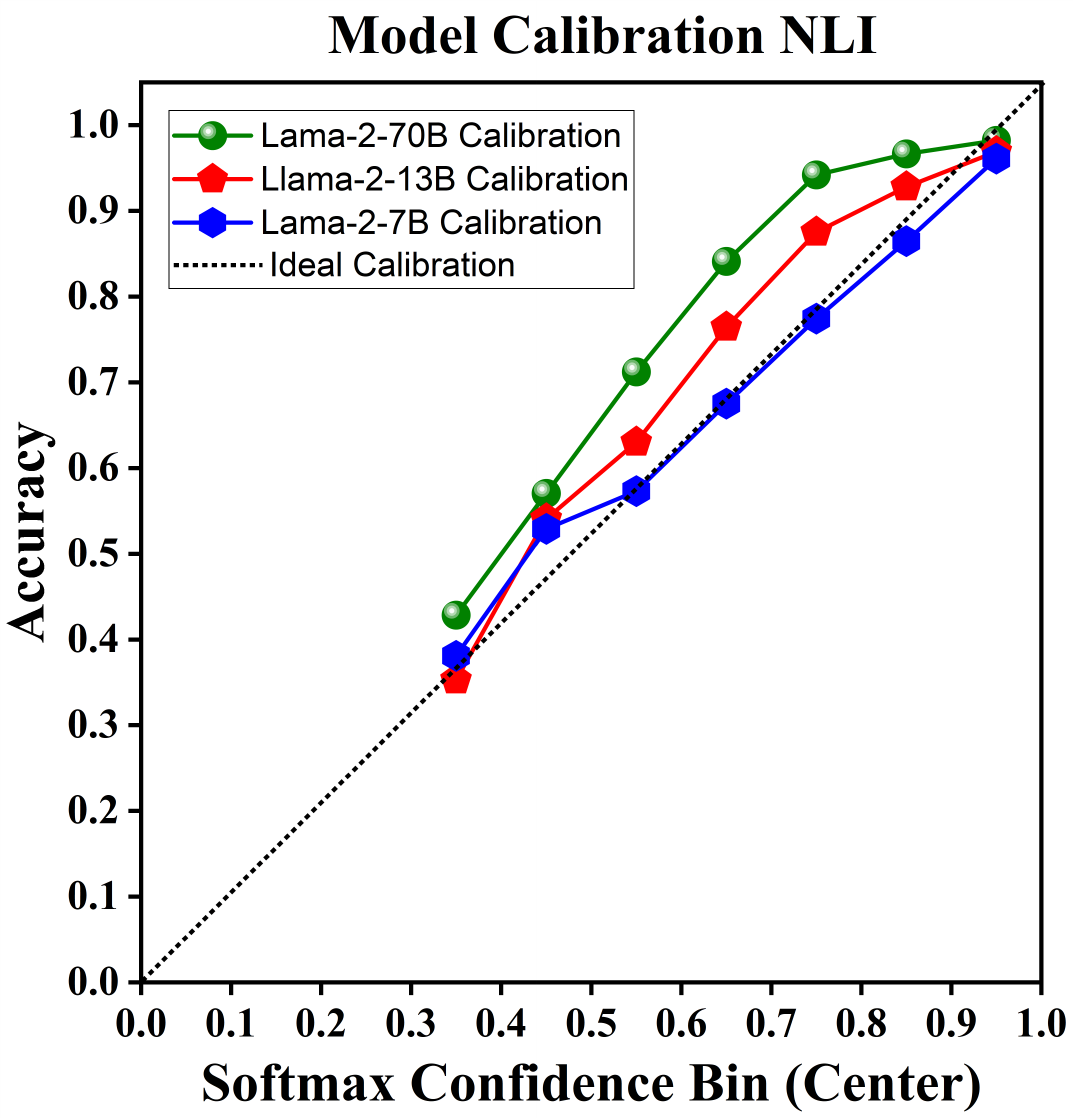}
\end{minipage}
\hfill
\begin{minipage}[t]{0.48\columnwidth}
\centering
\includegraphics[width=\linewidth]{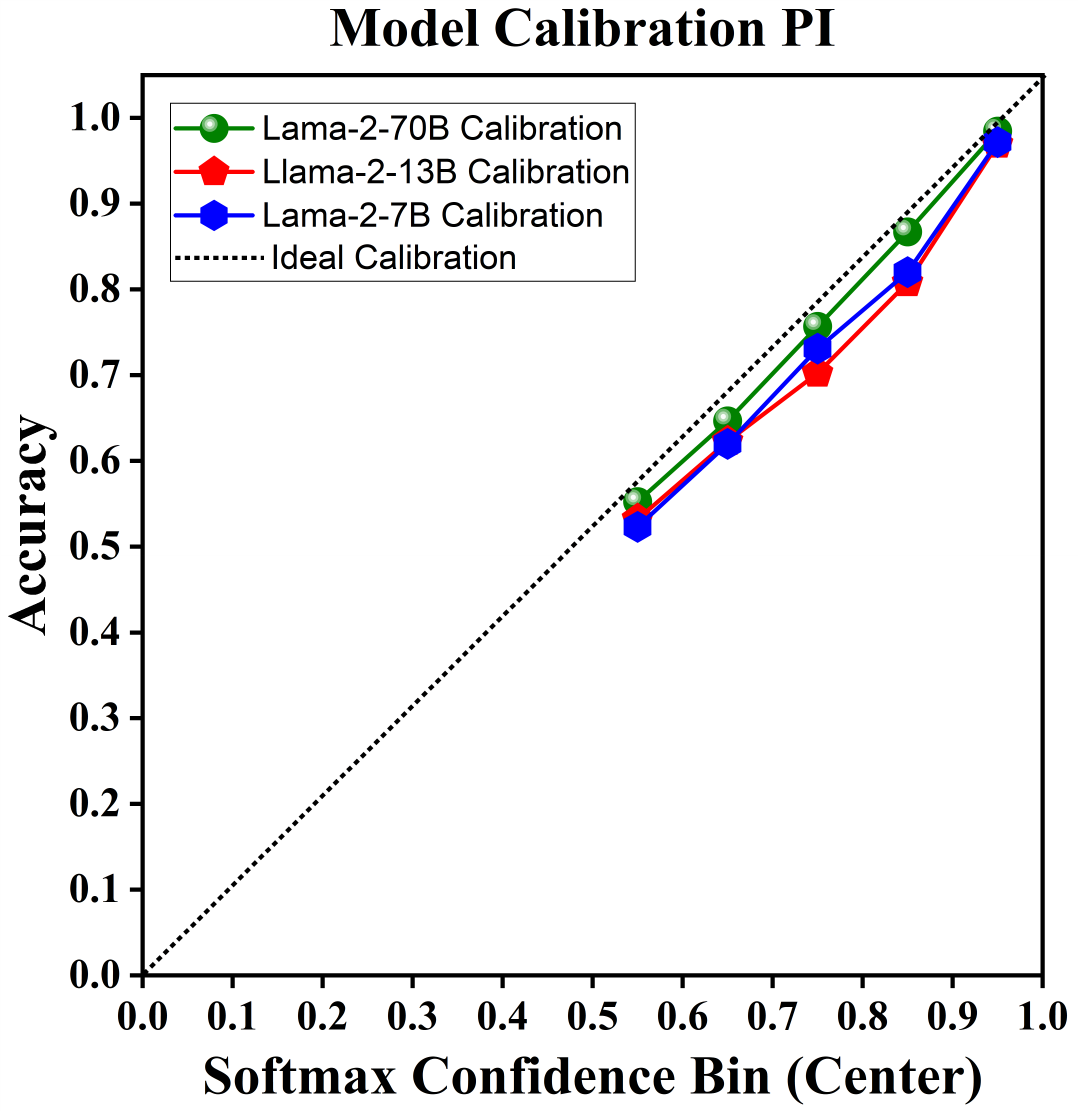}
\end{minipage}

\caption{Reliability diagrams for NLI and PI tasks for the Llama-2 family of models.}
\label{fig:model_reliability_llama}
\end{figure}

\begin{table}[hbt]
\renewcommand{\arraystretch}{1.1}
\caption{Softmax (S) and verbalized (V) confidence distributions for PI.}
\label{tab_llama_calibration_bins_pi}
\centering
\scriptsize
\setlength{\tabcolsep}{2pt}

\begin{tabular}{lcccccc}
\toprule
\textbf{Bin}
& \multicolumn{2}{c}{\textbf{70B}}
& \multicolumn{2}{c}{\textbf{13B}}
& \multicolumn{2}{c}{\textbf{7B}} \\
\cmidrule(lr){2-3}
\cmidrule(lr){4-5}
\cmidrule(lr){6-7}
& \textbf{S} & \textbf{V}
& \textbf{S} & \textbf{V}
& \textbf{S} & \textbf{V} \\
\midrule
0.0--0.1 & 0.00 & 0.00 & 0.00 & 0.00 & 0.00 & 0.00 \\
0.1--0.2 & 0.00 & 0.00 & 0.00 & 0.00 & 0.00 & 0.00 \\
0.2--0.3 & 0.00 & 0.00 & 0.00 & 0.00 & 0.00 & 0.00 \\
0.3--0.4 & 0.00 & 0.00 & 0.00 & 0.00 & 0.00 & 0.00 \\
0.4--0.5 & 0.00 & 0.00 & 0.00 & 0.00 & 0.00 & 0.01 \\
0.5--0.6 & 4.04 & 0.00 & 4.84 & 0.00 & 5.27 & 0.00 \\
0.6--0.7 & 4.64 & 0.00 & 5.71 & 0.00 & 6.12 & 0.00 \\
0.7--0.8 & 6.06 & 1.10 & 7.01 & 0.00 & 7.50 & 0.00 \\
0.8--0.9 & 10.35 & 30.89 & 12.12 & 0.00 & 12.20 & 0.00 \\
0.9--1.0 & 74.91 & 68.01 & 70.31 & 100.00 & 68.91 & 99.99 \\
\midrule
\textbf{Avg.}
& \textbf{92.08} & \textbf{93.78}
& \textbf{90.74} & \textbf{98.84}
& \textbf{90.38} & \textbf{99.00} \\
\bottomrule
\end{tabular}
\end{table}

\begin{table}[hbt]
\renewcommand{\arraystretch}{1.1}
\caption{Calibration similarity measurements for the Llama-2 family.}
\label{tab_ece_results_llama}
\centering
\scriptsize
\setlength{\tabcolsep}{2.5pt}

\begin{tabular}{lcccccc}
\toprule
\multirow{2}{*}{\textbf{Model}} &
\multicolumn{3}{c}{\textbf{NLI}} &
\multicolumn{3}{c}{\textbf{PI}} \\
\cmidrule(lr){2-4}
\cmidrule(lr){5-7}
& \textbf{ECE} & \textbf{MCE} & \textbf{Brier}
& \textbf{ECE} & \textbf{MCE} & \textbf{Brier} \\
\midrule
70B & 0.110 & 0.190 & 0.170 & 0.005 & 0.011 & 0.110 \\
13B & 0.055 & 0.120 & 0.190 & 0.017 & 0.050 & 0.160 \\
7B  & 0.013 & 0.065 & 0.220 & 0.014 & 0.035 & 0.160 \\
\bottomrule
\end{tabular}
\end{table}

\begin{table}[hbt]
\caption{Percentage change in calibration metrics of a smaller model ($M_2$) relative to a larger reference model ($M_1$) for Llama-2 family. Positive values indicate $M_2$ is worse than $M_1$; negative values indicate improvement.}
\label{tab_calibration_pct_llama}
\centering
\scriptsize
\setlength{\tabcolsep}{4pt}
\renewcommand{\arraystretch}{1.15}
\begin{tabular}{llrrrrrr}
\toprule
\multirow{2}{*}{$M_1$} & 
\multirow{2}{*}{$M_2$} & 
\multicolumn{3}{c}{\textbf{NLI}} & 
\multicolumn{3}{c}{\textbf{PI}} \\
\cmidrule(lr){3-5}\cmidrule(lr){6-8}
& & \textbf{ECE} & \textbf{MCE} & \textbf{Brier} & \textbf{ECE} & \textbf{MCE} & \textbf{Brier} \\
\midrule
70B & 13B & $-$50\% & $-$37\% & +12\%  & +240\%$^{*}$ & +355\%$^{*}$ & +45\% \\
70B & 7B  & $-$88\% & $-$66\% & +29\%  & +180\%$^{*}$ & +218\%$^{*}$ & +45\% \\
13B & 7B  & $-$76\% & $-$46\% & +16\%  & $-$18\%      & $-$30\%      & 0\% \\
\bottomrule
\end{tabular}
\vspace{1mm}
\begin{minipage}{\columnwidth}
\footnotesize
$^{*}$Large percentage swings in PI ECE and MCE reflect the near-perfect calibration of Llama-2-70B on this task (ECE = 0.005, MCE = 0.011), where even small absolute deviations produce large relative changes.
\end{minipage}
\end{table}

\subsection{Tasks and Datasets}

We evaluate trust-equivalence on two text classification tasks. 
\textbf{Natural Language Inference (NLI).} Given a premise and a hypothesis, the model must determine whether the hypothesis logically follows from, contradicts, or remains neutral with respect to the premise. This task tests relational and logical reasoning. We use the Stanford Natural Language Inference (SNLI) dataset~\cite{bowman2015large}. The training partition has approximately 500k pairs. Both validation and test partitions have approximately 10k pairs each.

\textbf{Paraphrase Identification (PI).} Given two questions, the model must determine whether they convey the same meaning. This task tests semantic understanding under lexical and syntactic variation. PI tests semantic equivalence. We use the Quora Question Pairs (QQP) dataset.\footnote{\url{https://www.kaggle.com/datasets/quora/question-pairs-dataset}} It contains over 400k question pairs labeled for semantic equivalence. The official QQP test set does not have separate train, validation and test partitions. So following the standard practice, we created our own splits from the official training data: 80\% for training, 10\% for validation, and 10\% for testing.

Together these two tasks form complementary stress tests. NLI requires reasoning about logical relationships between two sentences. Here key signals are often concentrated in specific relational cues such as negation and contradiction markers. PI requires judging overall semantic equivalence. Here key signals are distributed across the meaning of both questions. This difference in reasoning type, combined with the three-class versus binary label structure, ensures that our findings are not an artifact of any single task design.

Llama-2 is a decoder-only model. It is mainly used for generative tasks. We restrict our evaluation to discriminative classification (NLI and PI). This restriction is a methodological necessity. It is required to measure attribution alignment. LIME and SHAP need a fixed output space. They also need a shared scalar target to attribute against. In classification this is the class logit. These let us compute meaningful Jaccard similarities over input features. Open-ended generation does not provide a shared target. The models produce different text of different lengths. A clean 1:1 comparison of input attributions is then not well-defined. Classification gives us a controlled setting. It lets us isolate and measure the trust-equivalence gap. The results are not confounded by generative drift.

\subsection{Training Details}

All models were fine-tuned using the Hugging Face Transformers library~\cite{wolf-etal-2020-transformers} in PyTorch on a single NVIDIA A100-SXM4 GPU with 80GB of memory. We used QLoRA~\cite{dettmers2023qlora} with 4-bit NormalFloat (NF4) quantization. It allows all three model variants to fit within the GPU memory during fine-tuning. We used an effective batch size of 64 and the Paged AdamW 32-bit optimizer with a learning rate of $2\times10^{-4}$. A cosine learning rate schedule was applied throughout.

\section{Experimental Results}
We evaluate trust-equivalence along two dimensions: attribution alignment and calibration similarity. Our central claim is that performance parity does not guarantee trust-equivalence. The results presented here test this claim systematically across
two tasks and multiple evaluation tools. The findings are consistent. Smaller Llama-2 models do not maintain trust-equivalence with the 70B model. But the nature of the
gap depends on multiple factors.

\subsection{Attribution Alignment}

We measure attribution alignment using LIME~\cite{ribeiro-etal-2016-trust} and SHAP~\cite{NIPS2017_8a20a862}. Both have known instabilities~\cite{alvarez2018robustness}. However, they remain the most widely used model-agnostic explainability methods and provide consistent insights. We ran attribution alignment tests three times. All results are mean values of those three runs. Variance across runs is close to zero.

For each test instance, we extract the top-$K$ most influential words for each model. We treat the top-$K$ features as an unordered set to avoid sensitivity to small ranking variations. We then compute the Jaccard similarity coefficient between the feature sets of two models. Instance-level scores are averaged across the test set to produce a single alignment score for each model pair. Here we report results for $K=3$. 

The average input length for both tasks is approximately 25 tokens. We therefore varied $K$ from 1 to 5, corresponding to at most 20\% of the average input length. Alignment scores remain consistently low across all values of $K$ for both LIME and SHAP on both tasks. Our findings are not an artifact of the choice of $K$. They reflect a genuine and robust difference in the feature attribution of larger and smaller models.

Please refer to Tables~\ref{tab_example_alignment_llama_NLI} and~\ref{tab_example_alignment_llama_PI} for qualitative examples. They show the top-3 words highlighted by LIME for each model on a single NLI and PI test instance respectively. All models predict the correct label. But they focus on strikingly different words. Smaller Llama-2 models do not always align with the 70B model in their feature selection. Scaling down a model leads to a different way of processing the same input.

Please refer to Table~\ref{tab_alignment_results_llama} for
quantitative results with $K$=3. SHAP values show higher agreement between model pairs than LIME values across both tasks. 



\subsection{Calibration Similarity}

We measure calibration similarity by comparing confidence distributions, reliability diagrams, and three complementary
metrics across model pairs. We use the maximum softmax probability as the confidence
score for each prediction. It is the highest class probability output by the model's final layer. We also collect verbalized confidence. For this, we prompt each model to explicitly state its confidence as a percentage alongside its prediction. Confidence distributions are reported in
Tables~\ref{tab_llama_nli_calibration_bins}
and~\ref{tab_llama_calibration_bins_pi}. Reliability diagrams are shown in Figure~\ref{fig:model_reliability_llama}.
Quantitative calibration scores are reported in Table~\ref{tab_ece_results_llama}. Pairwise percentage changes in calibration scores are reported in
Table~\ref{tab_calibration_pct_llama}.

The most striking finding from the confidence distribution
analysis is that smaller models are severely overconfident
in their verbalized self-assessments while the 70B is not.
We compare softmax-based confidence with verbalized
confidence by prompting each model to generate its confidence
score alongside its prediction. The 70B remains relatively
underconfident in its verbalized scores. The 13B and 7B
report average verbalized confidence of approximately 99\%
on both tasks. This is a sharp divergence from their
softmax-based profiles. Smaller models cannot reliably
assess their own uncertainty. This is a distinct and
practically important form of trust-equivalence failure
that goes beyond miscalibration in the conventional sense.

Reliability diagrams reveal that the direction of
miscalibration reverses across tasks. For NLI, all models
are underconfident. Their accuracy curves lie above the
ideal calibration line. The 7B is closest to the diagonal.
For PI, all models are overconfident. Their accuracy curves
fall below the diagonal. The calibration profile is not
stable across tasks. A user who characterizes a model's
calibration behavior on one task cannot assume the same
behavior on another.

Quantitative metrics confirm that there is no monotone
relationship between model size and calibration quality.
For NLI, ECE and MCE decrease as model size decreases.
The 7B achieves the lowest ECE and MCE. This aligns with
the reliability diagram where the 7B curve lies closest
to the diagonal. However, the Brier Score increases as
model size decreases. Smaller models assign lower
probability to correct predictions, degrading sharpness.
For PI, the 70B achieves the lowest values across all
three metrics. The direction of the size-calibration
relationship depends on the task and the metric.

Pairwise percentage change analysis shows that the dominant
calibration gap is between the 70B and the smaller variants;
within the smaller models the gap is substantially smaller.
On NLI, the 7B shows 88\% lower ECE and 66\% lower MCE
than the 70B, but 29\% worse Brier Score. The 13B shows
a similar but less extreme pattern. On PI, both smaller
models show dramatically worse ECE and MCE relative to the
70B (+240\% and +355\% for the 13B; +180\% and +218\% for
the 7B), reflecting the near-perfect calibration the 70B
has acquired on this task (ECE\,=\,0.005, MCE\,=\,0.011)
that smaller models do not inherit. However, when the 13B
and 7B are compared directly, the gaps are much smaller.
On PI, the 7B actually improves over the 13B on ECE and
MCE ($-$18\% and $-$30\%) with identical Brier Score (0\%).
The dominant trust-equivalence failure is between the 70B
and the rest, not within the smaller models themselves.

A deployment engineer relying on any single metric or any
single task would draw different and potentially
contradictory conclusions about trust-equivalence. ECE and
MCE suggest smaller models are better calibrated on NLI
but worse on PI. The Brier Score suggests they are worse
on both. Verbalized confidence suggests they cannot assess
their own uncertainty at all. The full picture requires
all three metrics, both tasks, and explicit pairwise
comparison. Trust-equivalence on calibration cannot be
assumed and cannot be assessed from any single number.

\subsection{Results on BERT and ViT Families}

We validate our trust-equivalence framework on two additional
transformer families to test whether our findings generalize
beyond decoder-only LLMs. For BERT, we evaluate BERT-base
against four smaller variants: DistilBERT, BERT-Medium,
BERT-Mini, and BERT-Tiny. For ViT, we evaluate three
independently trained variants: ViT-Base, ViT-Large, and
ViT-Huge on ImageNet and CIFAR-100. Attribution alignment
for ViT is measured using Integrated Gradients and Occlusion
tests. ViT models process images as sequences of patches.
A standard 224×224 image with 16×16 patches produces a
sequence of 196 patches. We use $K$=10, which corresponds
to approximately 5\% of the patch sequence. This is
proportionally comparable to $K$=3 for text inputs of
25 tokens.

Attribution alignment is low across both families
(Tables~\ref{tab_alignment_results}
and~\ref{tab_alignment_results_vit}). BERT models disagree
on which input words matter most. ViT models disagree on
which image patches are most important, even when all
variants correctly classify the same image. ViT Jaccard
scores are the lowest across all three families, with no
pair exceeding 0.21 even at $K$=10. An important difference
emerges in the LIME-SHAP relationship. For BERT, LIME
scores are consistently higher than SHAP scores. This is
the opposite of what we observe for Llama-2. The
relationship between attribution methods and model
architecture is not fixed. It depends on the family. This
reinforces the importance of using both methods rather than
relying on either alone.

In the BERT family, the opposing pattern between ECE/MCE
and Brier Score holds
(Table~\ref{tab_calibration_pct}). Smaller BERT models
achieve lower ECE and MCE than BERT-base in most cases.
But their Brier Scores are uniformly worse. The pattern
strengthens with model size reduction. BERT-Tiny shows
68\% lower ECE and 89\% lower MCE on NLI, but 81\% worse
Brier Score. Apparent improvement on average and worst-case
calibration masks a consistent degradation in sharpness.

In the ViT family, calibration profiles diverge in a more
complex pattern driven by ViT-Huge's poor calibration
(Table~\ref{tab_calibration_pct_vit}). Both ViT-Large and
ViT-Base are substantially better calibrated than ViT-Huge
across all metrics and both datasets. However, when
ViT-Large and ViT-Base are compared directly, ViT-Base is
dramatically worse. On ImageNet, ViT-Base shows 323\%
higher ECE than ViT-Large. On CIFAR-100, the gap reaches
568\%. There is no monotone relationship between model size
and calibration quality. The direction of the gap depends
entirely on which pair is being compared.

Trust-equivalence failure is not specific to any family,
modality, or training paradigm. It holds for encoder-only
models. It holds for vision transformers. The nature of
the failure differs across families. The LIME-SHAP
relationship reverses between BERT and Llama-2.
Calibration patterns shift across families. But the core
finding does not change. The trust gap between models in
the same family must be measured explicitly, regardless
of which family is being deployed.

\begin{table}[t]
\caption{Attribution Alignment between BERT models using Jaccard coefficient ($K=3$).}
\label{tab_alignment_results}
\centering
\scriptsize
\setlength{\tabcolsep}{3pt}
\renewcommand{\arraystretch}{1.1}

\begin{tabular}{llcccc}
\toprule
\textbf{$M_1$} &
\textbf{$M_2$} &
\multicolumn{2}{c}{\textbf{NLI}} &
\multicolumn{2}{c}{\textbf{PI}} \\
\cmidrule(lr){3-4}
\cmidrule(lr){5-6}
& & \textbf{LIME} & \textbf{SHAP} & \textbf{LIME} & \textbf{SHAP} \\
\midrule

BERT-Base   & Distil-BERT & 0.45 & 0.36 & 0.52 & 0.3 \\
BERT-Base   & BERT-Medium & 0.56 & 0.44 & 0.54 & 0.21 \\
BERT-Base   & BERT-Mini   & 0.49 & 0.41 & 0.51 & 0.16 \\
BERT-Base   & BERT-Tiny   & 0.44 & 0.49 & 0.51 & 0.08 \\
\bottomrule

\end{tabular}
\end{table}

\begin{table}[t]
\caption{Attribution alignment between Vision Transformer models using Jaccard coefficient ($K=10$).}
\label{tab_alignment_results_vit}
\centering
\scriptsize
\setlength{\tabcolsep}{3pt}
\renewcommand{\arraystretch}{1.1}

\begin{tabular}{llcccc}
\toprule
\multirow{2}{*}{$M_1$} &
\multirow{2}{*}{$M_2$} &
\multicolumn{2}{c}{\textbf{ImageNet}} &
\multicolumn{2}{c}{\textbf{CIFAR-100}} \\
\cmidrule(lr){3-4}
\cmidrule(lr){5-6}
& &
\textbf{Occlusion} &
\textbf{Inte. Gradients} &
\textbf{Occlusion} &
\textbf{Inte. Gradients} \\
\midrule

ViT-huge  & ViT-large & 0.0628 & 0.1240 & 0.0621 & 0.1206 \\
ViT-huge  & ViT-base  & 0.0701 & 0.1370 & 0.0735 & 0.1347 \\
ViT-large & ViT-base  & 0.1013 & 0.2060 & 0.0922 & 0.1796 \\

\bottomrule
\end{tabular}
\end{table}

\begin{table}[hbt]
\caption{Percentage change in calibration metrics relative to BERT-base ($M_1$).}
\label{tab_calibration_pct}
\centering
\scriptsize
\setlength{\tabcolsep}{3pt}
\renewcommand{\arraystretch}{1.15}

\begin{tabular}{lrrrrrr}
\toprule
\multirow{2}{*}{$M_2$} &
\multicolumn{3}{c}{\textbf{NLI}} &
\multicolumn{3}{c}{\textbf{PI}} \\
\cmidrule(lr){2-4}\cmidrule(lr){5-7}
& \textbf{ECE} & \textbf{MCE} & \textbf{Brier}
& \textbf{ECE} & \textbf{MCE} & \textbf{Brier} \\
\midrule

\textbf{Distil-BERT}
& +14\% & $-$21\% & +19\%
& 0\% & $-$8\% & +6\% \\

\textbf{BERT-Medium}
& $-$18\% & $-$53\% & +13\%
& $-$3\% & $-$25\% & +6\% \\

\textbf{BERT-Mini}
& $-$58\% & $-$82\% & +31\%
& $-$44\% & $-$63\% & +6\% \\

\textbf{BERT-Tiny}
& $-$68\% & $-$89\% & +81\%
& $-$66\% & $-$75\% & +38\% \\

\bottomrule
\end{tabular}
\end{table}

\begin{table}[hbt]
\caption{Percentage change in calibration metrics of a smaller model ($M_2$) relative to a larger reference model ($M_1$) in the Vision Transformer family.}
\label{tab_calibration_pct_vit}
\centering
\scriptsize
\setlength{\tabcolsep}{2pt}
\renewcommand{\arraystretch}{1.1}

\begin{tabular}{llrrrrrr}
\toprule
\multirow{2}{*}{$M_1$} &
\multirow{2}{*}{$M_2$} &
\multicolumn{3}{c}{\textbf{ImageNet}} &
\multicolumn{3}{c}{\textbf{CIFAR-100}} \\
\cmidrule(lr){3-5}
\cmidrule(lr){6-8}
& & ECE & MCE & Brier & ECE & MCE & Brier \\
\midrule

ViT-huge & ViT-large & $-$91\% & $-$48\% & $-$52\% & $-$86\% & $-$25\% & $-$52\% \\
ViT-huge & ViT-base & $-$63\% & $-$36\% & $-$43\% & $-$6\%  & 0\%     & $-$19\% \\
ViT-large & ViT-base & +323\%  & +24\%   & +17\%   & +568\% & +32\%   & +69\%   \\

\bottomrule
\end{tabular}
\end{table}
\section{Conclusion and Future Work}

A smaller LLM cannot be trusted in the same way as its
larger counterpart in the same family. This is our central
finding. It holds across both tasks, both evaluation
dimensions, and across the BERT and ViT families where we
additionally validated our framework.

The failure is systematic but not uniform. Attribution alignment is generally low across all three families. Smaller models base their predictions on different input features than the larger model. The relationship between attribution methods and model architecture is not fixed. LIME scores are higher than SHAP scores for BERT, but lower for Llama-2. Using both methods is necessary to get a complete picture. Calibration profiles differ substantially across sizes with no consistent relationship between size and calibration quality. 


Attribution alignment and calibration similarity are not
the only dimensions of trust-equivalence. Robustness to
adversarial inputs~\cite{ye2019adversarial}, fairness across
demographic groups~\cite{hooker2019compressed}, and
vulnerability to membership inference
attacks~\cite{yuan2022membership} are all relevant. Our
framework is open-ended by design. The deployment context
should drive the choice of dimensions.


Several directions remain open for future work. Generation
tasks and open-ended settings are a natural next step beyond
classification. Training methods that explicitly preserve
trust-equivalence across sizes are an important open problem.
Extending the framework to cover robustness, fairness, and
privacy would make it more comprehensive. A mechanistic
explanation of why trust gaps arise within a family remains
an open question. Understanding the cause is the first step
toward designing solutions.

\bibliographystyle{IEEEtran}
\bibliography{references}

\end{document}